\documentclass[conference]{IEEEtran}
\IEEEoverridecommandlockouts
\usepackage{cite}
\usepackage{amsmath,amssymb,amsfonts}
\usepackage{algorithmic}
\usepackage{listings}
\usepackage{graphicx}
\usepackage{textcomp}
\usepackage{mdframed}
\usepackage{ulem}
\usepackage{colortbl}
\usepackage{hyperref}
\hypersetup{
    colorlinks=true,
    citecolor=blue
}

\lstdefinestyle{mystyle}{
    language=Python,
    basicstyle=\ttfamily,
    keywordstyle=\color{blue},
    stringstyle=\color{red},
    commentstyle=\color{green},
    morekeywords={as, self, pd, Union},  
    numbers=left,
    numberstyle=\tiny\color{gray},
    numbersep=5pt,
    showstringspaces=false
}

\usepackage[usenames,dvipsnames]{xcolor}
\def\BibTeX{{\rm B\kern-.05em{\sc i\kern-.025em b}\kern-.08em
    T\kern-.1667em\lower.7ex\hbox{E}\kern-.125emX}}
\begin{document}

\title{
A Comprehensive Python Library for Deep Learning-Based Event Detection in Multivariate Time Series Data \\ and Information Retrieval in NLP
}
\author{\IEEEauthorblockN{ Menouar Azib}
\and
\IEEEauthorblockN{ Benjamin Renard}
\and
\IEEEauthorblockN{ Philippe Garnier}
\and
\IEEEauthorblockN{ Vincent Génot}
\and
\IEEEauthorblockN{ Nicolas André}
}

\definecolor{RoyalBlue}{rgb}{0.25, 0.41, 0.88}
\definecolor{Peach}{rgb}{1.0, 0.9, 0.71}
\definecolor{Violet}{rgb}{0.56, 0.0, 1.0}
\definecolor{YellowGreen}{rgb}{0.6, 0.8, 0.2}
\definecolor{RedOrange}{rgb}{1.0, 0.27, 0.0}

\maketitle

\begin{abstract}
Event detection in time series data is crucial in various domains, including finance, healthcare, cybersecurity, and science. Accurately identifying events in time series data is vital for making informed decisions, detecting anomalies, and predicting future trends. Despite extensive research exploring diverse methods for event detection in time series, with deep learning approaches being among the most advanced, there is still room for improvement and innovation in this field. In this paper, we present a new deep learning supervised method for detecting events in multivariate time series data. Our method combines four distinct novelties compared to existing deep-learning supervised methods. Firstly, it is based on regression instead of binary classification. Secondly, it does not require labeled datasets where each point is labeled; instead, it only requires reference events defined as time points or intervals of time. Thirdly, it is designed to be robust by using a stacked ensemble learning meta-model that combines deep learning models, ranging from classic feed-forward neural networks (FFNs) to state-of-the-art architectures like transformers. This ensemble approach can mitigate individual model weaknesses and biases, resulting in more robust predictions. Finally, to facilitate practical implementation, we have developed a Python package to accompany our proposed method. The package, called eventdetector-ts, can be installed through the Python Package Index (PyPI). In this paper, we present our method and provide a comprehensive guide on the usage of the package. We showcase its versatility and effectiveness through different real-world use cases from natural language processing (NLP) to financial security domains.
\end{abstract}


\section{Introduction}
In recent years, deep learning techniques have shown promising results in various domains, including event detection in time series data. Notably, \cite{b1, b2, b3} provide comprehensive reviews of state-of-the-art deep learning and graph-based techniques for time series anomaly detection, respectively. Additionally, to the best of our knowledge, the most comprehensive anomaly detection benchmarks are provided by \cite{b4, b5}. These works demonstrate the potential of advanced machine learning techniques for detecting events in time series data. 

Unlike traditional binary classification methods, where each time step is labeled as an event or non-event, our approach is based on regression. Regression enables the prediction of continuous values, offering more nuanced information about event presence and strength in the time series. This approach is advantageous for complex and noisy data, where events may be less defined or easily distinguished from non-events. The regression approach can predict a continuous output variable, which can be beneficial when dealing with events that vary in intensity or magnitude. This flexibility allows the model to capture more detailed information about the event, such as its duration or severity. In real-world data, events often occur amidst background noise. Binary classification might struggle with this, as it requires a clear distinction between event and non-event classes. Regression, on the other hand, can handle this complexity better by predicting a continuous spectrum of values. The output of a regression model can provide more interpretable results. For instance, a higher predicted value could indicate a stronger or more significant event. This can provide more precise and actionable insights for decision-making. In many real-world scenarios, events of interest are rare compared to non-events. This leads to imbalanced data, which can pose challenges for binary classification models. Regression models can potentially handle such scenarios better by focusing on predicting the event’s characteristics rather than just its occurrence. Using regression, our method provides more precise and accurate insights into event characteristics within the time series data.

Secondly, our method does not require labeling each time step within a dataset, which can be a time-consuming and labor-intensive process. Instead, our method only requires reference events to be defined as time points or intervals of time, making it more efficient and practical to use. Thirdly, our method utilizes a stacked ensemble learning meta-model which allows for improved performance by leveraging the strengths of multiple base models and optimizing their predictions \cite{b6}. This approach enhances the overall accuracy and robustness of event detection.

Finally, to facilitate practical implementation, we have developed a Python package that accompanies our proposed method. This package provides an easy-to-use interface for applying our method to real-world datasets and allows users to quickly and easily detect events in their own time series data. We provide detailed documentation and examples to help users get started with our package and apply it to their own data. Overall, our method offers a powerful and flexible approach for detecting events in multivariate time series data that is both accurate and efficient.

The rest of this paper is organized as follows: In Section 2, we provide a detailed description of the regression aspect of our method, explaining how it differs from traditional binary classification approaches. In Section 3, we present our stacked ensemble learning approach, which combines the strengths of multiple base models to improve the accuracy and robustness of event detection. In Section 4, we describe the design and implementation of the Python package, which provides an easy-to-use interface for applying our method to real-world datasets. In Section 5, we showcase the usage of the package and demonstrate the effectiveness of the method through a series of experiments on different real-world datasets from NLP to financial security. Finally, in Section 6, we conclude the paper by summarizing our key findings and contributions. We highlight the strengths and limitations of our method and discuss potential future directions for research in this area.

\section{Regression Based Approach}
\subsection{Data}
In our approach, we require two pieces of data: one representing the time series (dataset) $S$, and the other representing the list of reference events (or ground truth events) $E$. The time series $S$ can be represented as follows:

\begin{equation}
\begin{aligned}
S: \mathbb{R} &\rightarrow  \mathbb{R}^f \\
t &\mapsto S(t)
\end{aligned}
\end{equation}

where $t$ represents the time index and $S(t)$ is a vector of $f$ features observed at time $t$. The list of reference events $E$ is represented either by a single time value or an interval. Each reference event in the list can be represented as a tuple $(t_{start},t_{end})$, where $t_{start}$ and $t_{end}$ are the start and end times of the event, respectively. If the event is represented by a single time value, then $t_{start} = t_{end}$. This list of reference events serves as the ground truth for evaluating the performance of the proposed event detection method.

Within the package, the time series $S$ and the events $E$ are typified as follows:

\begin{mdframed}[linecolor=black, topline=true, bottomline=true, leftline=false, rightline=false, backgroundcolor=yellow!20!white]
\begin{lstlisting}[language=Python]
import pandas as pd
from typing import Union
 # Time Series `S`
dataset: pd.DataFrame
events: Union[list, pd.DataFrame] 
\end{lstlisting}
\end{mdframed}

The \lstinline|dataset| variable represents the time series data in a Pandas DataFrame format, where the index represents the time and the columns represent the features ($f$). The \lstinline|events| variable represents the reference events in either a list or Pandas DataFrame format. They can have either one column to represent the midtime of events or two columns to represent the beginning and ending time of each event.

\subsection{Sliding Windows}
The proposed method is based on regression, so we must create a scalar parameter for this task based on the required time series $S$ and a list of reference events $E$. To achieve this, we use a sliding window approach to divide the time series data into overlapping windows. Each window is represented by a fixed-size ($width$) segment of observations, and the windows are shifted along the time axis by a fixed step size $step$. We associate each window with a real value called the overlapping parameter $op \in [0,1]$, which is computed as follows: 
\begin{equation}
op(w_i, e_j) = \dfrac{{\text{duration}(w_i \cap e_j)}}{{\text{duration}(w_i \cup e_j)}}
\end{equation}
where $w_i$ represents the $i$-th window in the time series, which is a matrix of observations spanning the duration of the window. Each row of the matrix corresponds to a time step within the window, and each column corresponds to one of the $f$ features observed at that time step. On the other hand, $e_j$ represents the $j$-th event in the list of reference events.

To facilitate the comparison between the reference events and the sliding windows, we convert the reference events into fixed-size intervals denoted by $width\_events$. This can be done by extending or truncating each reference event to match the desired interval size. The choice of interval size should be carefully considered, as it can affect the performance of the method. A larger interval size provides more flexibility in matching the reference events with the sliding windows, but it also increases the risk of false positives. On the other hand, a smaller interval size reduces the risk of false positives, but it also increases the risk of false negatives. In practice, the optimal interval size can be determined through experimentation on a validation set.

The overlapping parameter measures the degree of overlap between a window and an event. If the window $w_i$ and the event $e_j$ do not intersect, the overlapping parameter will be 0, indicating no relevance between them. On the other hand, when $w_i$ and $e_j$ overlap, $op$ will be greater than 0, with the maximum value reached when a perfect match is observed. The value 1 is reached in case of a perfect match and when the sliding windows and the event windows have the same width.

To associate a single value of $op$ to each sliding window $w_i$, we take the maximum of the overlapping values among all events:

\begin{equation}
op(w_i) = \max_{e_j \in E} op(w_i, e_j)
\end{equation}

\begin{figure*}[t]
\centerline{\includegraphics[width=0.78\textwidth]{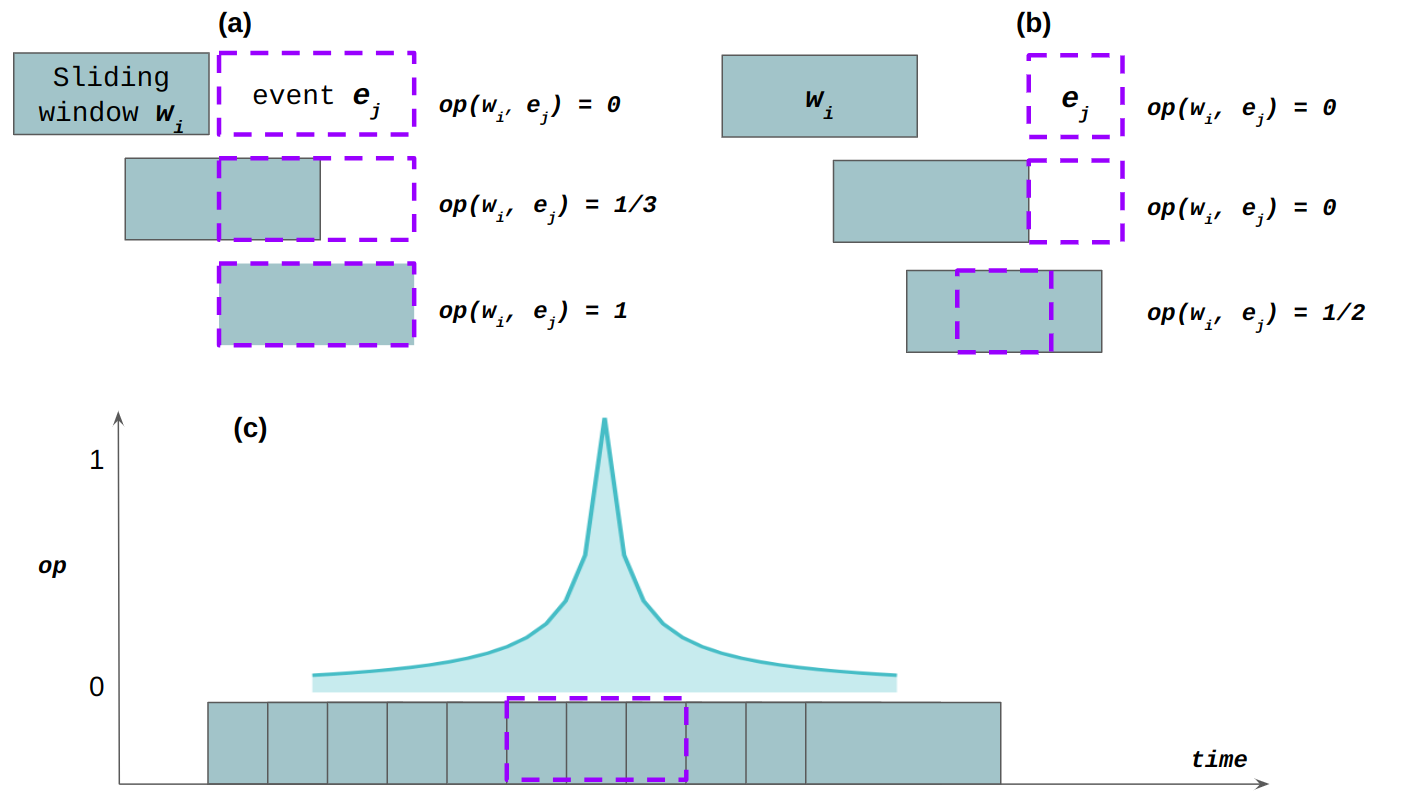}}
\caption{A visual illustration of the overlapping parameter $op$ between sliding windows $w_{i, i> 0}$ and an event $e_j$: (a) demonstrates the scenario when they have the same width, (b) depicts the case when the event is half the width of $w_i$, and (c) illustrates that the peak of the overlapping parameter $op$ is achieved at the midpoint of the relevant event associated with windows $w_i$}
\label{fig_1}
\end{figure*}

This means that for each sliding window, we select the event with the highest degree of overlap as the most relevant event. The overlapping parameter $op(w_i)$ provides a measure of how well the window aligns with the most relevant event, which can be used to guide a regression model in predicting the presence or absence of events in the time series data. Fig.~\ref{fig_1} illustrates this concept with a visual representation of the sliding windows and their corresponding overlapping parameters.

\subsection{Principle of Event Detection}
The principle of event detection using sliding windows and the overlapping parameter $op$ is rooted in training a deep learning model $M$. This model learns to predict the $op$ value for unseen windows $w_{p, p >0}$. Once trained, model $M$ can be applied to any window $w_p$ to obtain a predicted value of $op_p$. These predicted $op_p$ values can then be used to identify features indicative of events. Typically, the maximum values (peaks) in $op_p$ correspond to the mid-times of potential events, as illustrated in Fig.~\ref{fig_1}.

These peaks can be identified and analyzed to extract the events represented by the following intervals: \[ e_q = \left[t_q - \dfrac{width\_events}{2}, t_q + \dfrac{width\_events}{2}\right]\]
where $t_q$ represents the mid-time of the q-th peak and $width\_events$ is the width of events. The extracted events can then be compared with ground truth events to evaluate the model's performance. For a theoretical discussion on the existence of model $M$, based on \uline{universal approximation theorems}, please refer to \cite{b7}.

\section{Stacked Ensemble Learning}
To enhance the robustness of predicting $op$, we employ a stacked ensemble learning approach. This approach combines multiple deep learning models to improve the overall performance of the system. The concept behind stacked ensemble learning is to use the predictions of several base models as input to a meta-model, which then makes the final prediction. This allows the system to leverage the strengths of multiple models and reduce the impact of their individual weaknesses. By combining different models, the stacked ensemble can achieve better performance and improved robustness compared to using a single model alone. For more details, please refer to \cite{b7}.

\section{The Package}
The section of most interest to practitioners is where we present the package and explain how all the parameters we have discussed can be used and initialized. The package can be found \href{https://pypi.org/project/eventdetector-ts/}{here}.

\subsection{MetaModel}
The first thing to do when using this package is to instantiate the \lstinline|MetaModel| object. This object serves as the foundation for building and training the stacked ensemble meta-model. It provides a convenient interface for configuring the various parameters of the method and managing the training process. To initialize the \lstinline|MetaModel| object, we need to provide several arguments, including the time series dataset, the reference events, the sliding window width, the events width, and the output directory where the results and models will be saved.

The \texttt{MetaModel} class provides several optional arguments for customization. Comprehensive documentation for the \texttt{MetaModel} object and its methods, including a complete list of optional arguments, is available \href{https://github.com/menouarazib/eventdetector#documentation}{here}.

To initialize the meta-model, follow the steps outlined in the provided code snippet:
\begin{mdframed}[linecolor=black, topline=true, bottomline=true, leftline=false, rightline=false, backgroundcolor=yellow!20!white]
\begin{lstlisting}
from eventdetector_ts.metamodel.meta_model 
    import MetaModel
meta_model = MetaModel(
    dataset=dataset,    
    events=events,      
    width=width,       
    width_events=width_events,  
    output_dir=output_dir)
meta_model.prepare_data_and_computing_op()
meta_model.build_stacking_learning()
\end{lstlisting}
\end{mdframed}

\subsection{Event Extraction Optimization}
In practice, deep learning models used for regression tasks can sometimes generate noisy predictions. In our case, this noise can cause problems when estimating peak locations, leading to the identification of false events. To address this issue, we smooth the predictions using a Gaussian filter. However, to optimize the event extraction process, we need to fine-tune several parameters. These include the size of the Gaussian filter, its standard deviation, and the threshold used for peak consideration. By optimizing these parameters, we can more accurately identify predicted events and improve the overall performance of the meta-model. We evaluate the performance of this process using metrics such as the F1-Score, which combines precision and recall to provide a balanced evaluation of the algorithm’s ability to correctly identify events. The goal is to maximize the F1-Score by simultaneously optimizing both precision and recall. For more details, please refer to \cite{b7}. To apply this process and run the event extraction optimization, we can call the following method:
\begin{mdframed}[linecolor=black, topline=true, bottomline=true, leftline=false, rightline=false, backgroundcolor=yellow!20!white]
\begin{lstlisting}
meta_model.event_extraction_optimization()
\end{lstlisting}

\end{mdframed}

\subsection{Visualizing and Evaluating Results}
The final step is to visualize the results, including the losses of the stacked models and the meta-model, true/predicted $op$, true/predicted events, and a histogram illustrating the time shift between the true and predicted events. This offers a visual assessment of the meta-model's performance and an evaluation of its ability to predict the target events. We can make use of the \lstinline|plot_save| method provided by the package to generate and save these plots. To create and save these plots, we can utilize the following code snippet:

\begin{mdframed}[linecolor=black, topline=true, bottomline=true, leftline=false, rightline=false, backgroundcolor=yellow!20!white]
\begin{lstlisting}
meta_model.plot_save(show_plots=True)
\end{lstlisting}

\end{mdframed}

\section{Application Scenarios and Performance Evaluation}
In this section, our goal is to showcase the versatility and simplicity of our method by demonstrating its wide range of applications. We emphasize its effectiveness in delivering robust and accurate results for event detection in time series data, as well as its adaptability for information retrieval in NLP. To do this, we present five distinct cases. Four of these cases revolve around information retrieval in textual data within the NLP domain, while the remaining case addresses credit card fraud detection in the field of financial security. This diverse set of scenarios illustrates the flexibility and applicability of our method and underscores its usability across various domains.

\subsection{Datasets}
\subsubsection{Credit Card Fraud Detection}
The first case is focused on the detection of credit card fraud, utilizing a dataset provided by \cite{b8}. This dataset contains transaction data, encompassing both legitimate and fraudulent credit card transactions. Credit card fraud is a significant concern within the financial industry, involving the unauthorized use of someone else's credit card information for fraudulent purposes. The dataset represents a time series with a time sampling of 1 second, comprising 492 frauds (events) out of 284,807 transactions. Notably, this dataset exhibits a high level of class imbalance, with the positive class (frauds) accounting for only 0.172\% of all transactions. The initial dataset was made for binary classification, where a column represents fraud events. To be used by our package, we have converted this column into a list of reference events represented as time points based on the time sampling of the dataset.

\subsubsection{Information Retrieval in Textual Data for NLP}
In our second evaluation scenario, we shift our focus towards the extraction of \uline{keywords} and the identification of \uline{adjectives} for part of speech tagging (POS) in textual data. For detailed insights into keyword extraction, please consult \cite{b9, b10, b11, b12, b13, b14}. Additionally, for further information on part of speech tagging, refer to \cite{b15}. To conduct this evaluation, we draw upon two extensive English texts sourced from the Wikipedia dump \cite{b16}. One of these texts centers around Autism \cite{b17}, while the other delves into Anarchism \cite{b17}. To effectively utilize our package, originally tailored for real-time series data and temporal events, we must first convert these texts into real-time series. Subsequently, we'll establish a mapping between keywords and adjectives to represent them as real-time events.

To convert a text into a real-time series, we initially tokenize the text into individual words \cite{b18}. Following tokenization, we leverage word embedding techniques like Word2Vec \cite{b19, b20, b21}. Word2Vec transforms words into dense numerical representations within a high-dimensional space, typically comprising 300 features.

The code snippet illustrating this process can be found \href{https://github.com/menouarazib/InformationRetrievalInNLP#text-as-time-series-using-word2vec}{here}.

Given that keywords and adjectives may occur at various positions within the text (which has been transformed into a time series), we establish a mapping between these positions and corresponding timestamps based on the index of the obtained time series. This mapping enables us to convert occurrences of keywords and adjectives within the text into temporal values. Subsequently, by specifying a value for the event's width ($width\_events$), we represent them as temporal events. For this evaluation, we have selected a set of 20 reference keywords associated with the Autism and Anarchism texts. Additionally, the list of reference adjectives has been obtained using the Natural Language Toolkit library \cite{b18} on these texts. As a result, for each of these texts, we create two cases: one for keyword extraction (Autism (Keys) and Anarchism (Keys)) and another for finding adjectives (Autism (POS) and Anarchism (POS)), as outlined in TABLE.~\ref{tab1}.

\subsection{Results}
In this section, we present the result evaluations of our method across the specified use cases, assessing its performance using three key metrics: F1-Score, Precision, and Recall. The results of these evaluations are presented in TABLE.~\ref{tab1}. Each block represents a use case, and each line within the block is characterized by a distinct configuration of the meta-model, involving variations in sliding window width and stacked models. 

All the presented use cases with code can be found at the following links: \href{https://github.com/menouarazib/InformationRetrievalInNLP}{Information Retrieval in NLP} and \href{https://github.com/menouarazib/eventdetector#code-implementations}{Credit Card Fraud}.

\begin{table*}
\caption{Performance of Our Method Across Multiple Use Cases}
\begin{center}
\begin{tabular}{|c|c|c|c|c|c|c|c|c|c|c|}
\hline
\textbf{Dataset} & $width$ & $step$ & $width\_events$ & Stacked Models & Hyper-parameters & Meta-model & \textbf{F1-Score} & \textbf{Precision} & \textbf{Recall} \\

\arrayrulecolor{RoyalBlue}\hline
Credit card fraud & 2 & 1 & 1 & Default & Default & Average & 0.84 & 0.98 & 0.73 \\
 & \textcolor{blue}{3} & 1 & 1 & Default & Default & Average & 0.83 & 0.96 & 0.73 \\
 & \textcolor{blue}{2} & 1 & 1 & \textcolor{blue}{CNN, LSTM}  & Default & Average & 0.82 & 0.93 & 0.73 \\
& 2 & 1 & 1 & \textcolor{blue}{TRANSFORMER}  & Default & Average & 0.82 & 0.93 & 0.73 \\
& 2 & 1 & 1 & \textcolor{blue}{FFN}  & \textcolor{blue}{(1, 20, 20)} & Average & 0.84 & 1.0 & 0.73\\ 
\arrayrulecolor{Peach}\hline

Autism (Keys) & 2 & 1 & 1 & Default & Default & Average & 0.98 & 1.00 & 0.96 \\
& \textcolor{blue}{3} & 1 & 1 & Default & Default & Average & 0.97 & 0.99 & 0.96 \\
\arrayrulecolor{Violet}\hline

Autism (POS) & 2 & 1 & 1 & Default & Default & Average & 0.80 & 0.85 & 0.75 \\
& \textcolor{blue}{3} & 1 & 1 & Default & Default & Average & 0.80 & 0.85 & 0.75 \\
& 3 & 1 & 1 & \textcolor{blue}{GRU} & Default & Average & 0.80 & 0.80 & 0.80 \\
\arrayrulecolor{YellowGreen}\hline

Anarchism (Keys) & \textcolor{blue}{2} & 1 & 1 & Default & Default & Average & 0.99 & 1.00 & 0.98 \\
& 2 & 1 & 1 & \textcolor{blue}{CONV\_LSTM1D} & Default & Average & 0.98 & 0.98 & 0.98 \\

\arrayrulecolor{pink}\hline
Anarchism (POS) & 2 & 1 & 1 & Default & Default & Average & 0.77 & 0.89 & 0.68 \\
& \textcolor{blue}{3} & 1 & 1 & Default & Default & Average & 0.77 & 0.88 & 0.69 \\
& 3 & 1 & 1 & \textcolor{blue}{(CNN\_RNN, 2)} & Default & FFN & 0.79 & 1.00 & 0.65 \\
\arrayrulecolor{RedOrange}\hline

\hline
\end{tabular}
\label{tab1}
\end{center}
\end{table*}

\begin{figure*}[h]
\centerline{\includegraphics[width=\textwidth]{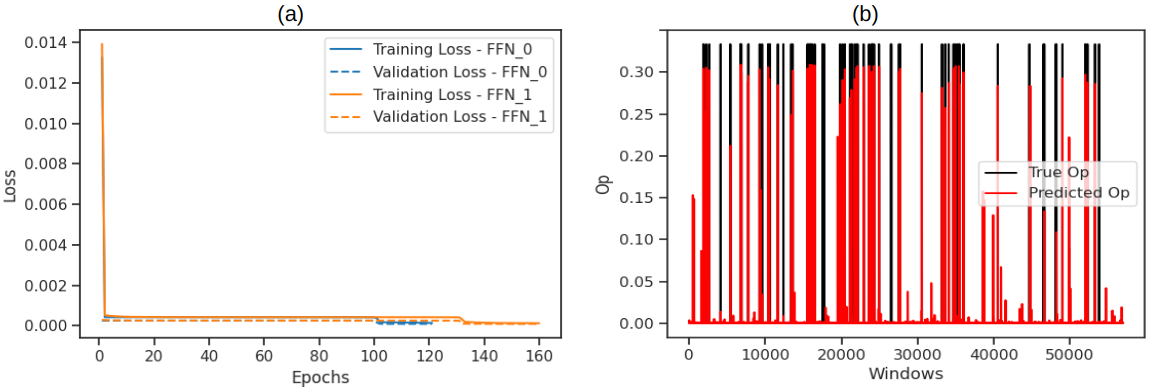}}
\caption{Training and validation losses of the stacked models (a), as well as the comparison of predicted and true $op$ (b) associated with the first row within TABLE.~\ref{tab1}.}
\label{fig_2}
\end{figure*}

Fig.~\ref{fig_2} displays the training and validation losses (MSE) of the stacked models during the training process for the first row within TABLE.~\ref{tab1}. The low losses observed in this case indicate that the meta-model has successfully learned the underlying patterns behind the $op$ values, justifying the good metrics obtained. Moreover, the predicted $op$ values exhibit a remarkable alignment with the true values, showing minimal fluctuations. This underscores the importance of employing a Gaussian filter to smoothen the predictions \cite{b7}. It's worth noting that occasional additional peaks are observed, emphasizing the need for employing a peak threshold to filter out false peaks. These steps ultimately contribute to enhancing the accuracy of the predictions. The results obtained showcase promising metrics, underscoring the effectiveness of our method across diverse domains. It is important to emphasize that our \textbf{primary objective} is not to attain state-of-the-art metrics for each use case presented here but to demonstrate our method's versatility and the easy use of the package in various domains. To achieve state-of-the-art metrics, a deeper exploration of hyper-parameter optimization for both the stacked models and the meta-model is necessary. This deeper optimization could lead to further enhancements in our method's performance, potentially yielding superior results. Even when employing the meta-model's default settings, we consistently observe robust metrics, as corroborated by a review analysis of credit card fraud detection performance \cite{b22}. The reported F1 scores, ranging from 0.80 to 0.89, align with our findings and affirm the reliability of our method.

\section{Conclusion}
We have introduced a novel deep-learning supervised method designed for the detection of events in multivariate time series data. Our method stands out due to several unique characteristics when compared to existing deep-learning supervised approaches. These include a focus on regression rather than binary classification, the capability to operate with a list of ground truth events as opposed to fully labeled datasets, the implementation of a stacked ensemble learning meta-model to enhance robustness, and the provision of a user-friendly Python package for practical implementations. Through a series of diverse use cases spanning natural language processing (NLP) tasks such as keyword extraction and part of speech tagging, as well as credit card fraud detection for financial security domains, we have underscored the universality of our method with a primary objective is not to attain state-of-the-art metrics rather than affirming its adaptability across various domains. The extensive and user-friendly API provided by our package simplifies the process of event detection for real-time data and other applications. It also allows for in-depth exploration of hyperparameter optimization for both the stacked models and the meta-model, offering the potential to achieve even more remarkable results with our method. However, it is essential to acknowledge that our method and the accompanying package have certain limitations. Notably, multi-class event detection may not be straightforward. In scenarios where multiple classes of events need to be detected, it becomes necessary to train separate meta-models to learn the $op$ associated with each individual class. While this approach adds complexity to the implementation, it also highlights an area for potential improvement in future iterations of our method. In addition, we may also want to test our approach on a variety of different datasets from different fields, with a focus on obtaining state-of-the-art metrics on each dataset compared to other methods. This would allow us to highlight the benefits of our method, such as the small number of parameters needed to train FFNs, which reduces the computational resources required and makes it a more sustainable and environmentally friendly solution. 

\section*{Acknowledgment}
This project was undertaken as part of the France Relance Program, under the agreement UT3 - N°0002890, with the support of the French Ministère de l’Economie des Finances et de la Souveraineté Industrielle et Numérique. We would like to express our gratitude to Akkodis, Université Paul Sabatier Toulouse III, CNRS, and IRAP. Their contributions have been invaluable to the success of this project.

\vspace{12pt}

\end{document}